\documentclass[sigconf]{acmart}

\usepackage{booktabs} 
\usepackage{amsmath}
\usepackage{algorithm}
\usepackage[noend]{algpseudocode}
\usepackage{multirow}

\copyrightyear{2021} 
\acmYear{2021} 
\setcopyright{acmcopyright}\acmConference[SAC '21]{The 36th ACM/SIGAPP Symposium on Applied Computing}{March 22--26, 2021}{Virtual Event, Republic of Korea}
\acmBooktitle{The 36th ACM/SIGAPP Symposium on Applied Computing (SAC '21), March 22--26, 2021, Virtual Event, Republic of Korea}
\acmPrice{15.00}
\acmDOI{10.1145/3412841.3441972}
\acmISBN{978-1-4503-8104-8/21/03}

\begin{document}
\title[Transfer Learning for Detecting Psychological Distress in Brexit Tweets]
{Transfer Learning Approach for Detecting Psychological Distress in Brexit Tweets}
\titlenote{Produces the permission block, and
  copyright information}


\author{Sean-Kelly Palicki}
\affiliation{%
\institution{School of Computing and Digital Technology}
  \institution{Birmingham City University}
  \state{United Kingdom} 
}

\author{Shereen Fouad}
\affiliation{%
\institution{School of Computing and Digital Technology}
  \institution{Birmingham City University}
  \state{United Kingdom} 
}

\author{Mariam Adedoyin-Olowe}
\affiliation{%
\institution{School of Computing and Digital Technology}
  \institution{Birmingham City University}
  \state{United Kingdom} 
}

\author{Zahraa S. Abdallah}
\affiliation{%
\institution{Department of Engineering Mathematics}
  \institution{University of Bristol}
  \state{United Kingdom} 
}

\renewcommand{\shortauthors}{S. Palicki et al.}

\begin{abstract}

 In 2016, United Kingdom (UK) citizens voted to leave the European Union (EU), which was officially implemented in 2020. During this period, UK residents experienced a great deal of uncertainty around the UK’s continued relationship with the EU. Many people have used social media platforms to express their emotions about this critical event. Sentiment analysis has been recently considered as an important tool for detecting mental well-being in Twitter contents. However, detecting the psychological distress status in political related tweets is a challenging task due to the lack of explicit sentences describing the depressive or anxiety status. To address this problem, this paper leverages a transfer learning approach for sentiment analysis to measure the non-clinical psychological distress status in Brexit tweets. The framework transfers the knowledge learnt from self-reported psychological distress tweets (source domain) to detect the distress status in Brexit tweets (target domain). The framework applies a domain adaptation technique to decrease the impact of negative transfer between source and target domains. The paper also introduces a Brexit distress index that can be used to detect levels of psychological distress of individuals in Brexit tweets. We design an experiment that includes data from both domains. The proposed model is able to detect the non-clinical psychological distress status in Brexit tweets with an accuracy of 66\% and 62\% on the source and target domains, respectively. 
\end{abstract}

%
%

\ccsdesc[500]{Applied computing~Psychology}
\ccsdesc[500]{Information systems~Sentiment analysis}
\ccsdesc[500]{Information systems~Social networking sites}
\ccsdesc[300]{Information systems~Data analytics}

\keywords{Transfer learning, Sentiment analysis, Brexit, Psychological distress, Social media analytics}

\maketitle

\section{Introduction}
\label{sec:intro}
On June 23rd, 2016 52\% of the people of the United Kingdom (UK) voted to leave the European Union (EU). The Brexit (``Britain Exit'') negotiation took almost 4 years until this separation was finally implemented in 2020, as a result of second referendum. During this period UK residence have experienced a great deal of uncertainty about the future relationship with the EU, and their social and economic stability across various sectors. This psychological distress is known as ``Brexit  Anxiety'' \cite{guma2019we}.
While the potential consequences of Brexit are often spoken of in terms of socio-economic impact, e.g. food-shortages and unemployment, there has also been an interest in understanding the potential psychological impact. For example,  
\cite{lansdall2016change} used Brexit as a natural experiment to study the effects of political and economic instability on psychological well-being, showing short-term changes in public mood such that the referendum triggered sadness, anxiety and anger. Previous research in \cite{vandoros2019eu} showed a measurable increase in the prescription of anti-depressant medication after Brexit. During the stressful period the UK residence have been using social media platforms to express their concerns and emotions about this major event.

Social media represents a wealth of information about public opinions. Recent years have witnessed rapid growth in the analysis of social media for studying a wide range of mental health problems \cite{yazdavar2020multimodal}. 
It is a multidisciplinary interest across linguistics, computer science, medicine, and psychology to use social media to automatically determine the mental state in populations \cite{morales2017cross, guntuku2017detecting}. In this era of research, there are two major approaches: the first focuses on using social media to model public mood as it relates to major public events. Changes in the emotional state have been correlated with major political events, natural disasters, cultural phenomenon, and in the prediction of stock prices \cite{bollen2011modeling, tumasjan2010predicting}. The second takes a clinical approach by using social media and machine learning techniques to classify abnormal psychological states (depression, anxiety, PTSD, etc.) in populations which had formerly been diagnosed through more traditional means \cite{de2013predicting, benton2017multi}. Our research exploits both techniques to understand the impact of major Brexit events on the mental well-being of UK residents using social media contents. We particularly focus on the final stages of Brexit negotiations during August/September 2019 when the nation was concerned over ``No-Deal Brexit''. 

Automatic detection of non-clinical depressive states (known here as psychological distress status) in social media political-related contents (tweets) is deemed a difficult task. This is because the political related tweets usually do not include clear posts in which users explicitly share their mental health problems, medical diagnosis or depressive thoughts. To address this problem, we apply a transfer learning approach for sentiment analysis where we train a supervised learning model on a large amount of labelled distress tweets and use it as a baseline to train Brexit tweets to detect the non-clinical psychological distress of individuals. Using the model, we introduce a Brexit Distress Index that may serve to characterise levels of psychological distress in populations in Brexit related tweets.


\section{Literature Review}
\label{sec:lit-review}
\subsection{Brexit and Psychological Well-Being}
 
The potential effects of Britain leaving the EU have been studied extensively, with an expected impact on areas such as free trade, immigration, social services, employment, and the role of EU law \cite{vandoros2019eu}. The result for UK residents, regardless of whether they voted for Brexit, is a high degree of political and economic uncertainty. Subsequently, research shows that such uncertainty is associated with higher rates of psychological distress, depression, and even suicide rates \cite{jofre2018impact}.

Previous research provides evidence that the early stages of the Brexit process impacted the psychological well-being of UK residents. \cite{vandoros2019eu} established a relationship between mental-health and Brexit by comparing change in prescriptions for antidepressant medications across England between January 2011 and December 2016. The study found a significant increase in antidepressant medications following the Brexit referendum. Notably, there were no significant differences when they compared the results between pro-remain and pro-leave dominant voting areas. Using Twitter, \cite{lansdall2016change} measured hourly sentiment expressed in the days before and after the Brexit Referendum. This analysis reported increases in negative affect, anger, anxiety, and sadness, with a corresponding decrease in positive affect.

This paper will demonstrate evidence for long-term psychological effects of the Brexit process. As mental health issues have been associated with complex language patterns which are not immediately detected by standard sentiment analysis libraries, we improved upon previous research by introducing transfer learning based sentiment analysis to identify psychological distress. This method to measure public mood may be used as the UK transitions out of the EU.  

\subsection{Automatically Detecting Mental Health Signals}
Automatically detecting psychological distress through social media could lead to improved psychological assessment, treatment, and support; especially for those who may have been otherwise unidentified. Social networks have been used to detect a list of conditions, including: Depression, PTSD, Anxiety, Bipolar Disorder, Seasonal Affective Disorder, Eating Disorder, Schizophrenia, Suicidal Ideation, Phobias, and others \cite{wongkoblap2017researching}. \cite{guntuku2017detecting} reviewed successful strategies for classifying mental health disorders and reported machine learning algorithms such as Support Vector Machine, Regression, and Random Forest can identify patterns shared by individuals with diagnosed mental health concerns. However, a significant challenge for this domain has been to obtain labelled mental health related data. Consequently, researchers have applied varied techniques to identify those with concerns, including: manual labelling by clinical mental health experts, getting consent to access patient health records, having individuals complete psychometric surveys, and self-declaration by users.

Using over 13 million tweets from individuals who explicitly stated they had been diagnosed with clinical depression or PTSD, \cite{coppersmith2014quantifying} distinguished between ‘diagnosed’ and 'non-diagnosed' participants, but not between Depression and PTSD, using language patterns. In a later paper, \cite{coppersmith2015clpsych} used a similar data collection strategy and to identify mental health concerns and found groups of words that classified between Depression, PTSD, and Control groups with 80-90\% precision. In \cite{de2013predicting} users completed a psychometric survey querying signs of depression and shared three months of Twitter posts. Researchers extracted features from the tweets including post sentiment, linguistic style, n-grams, proportion of posts with reply, and number of followers. They found depressed users had significantly less Twitter engagement, more emotional word related n-grams, negative affect, and less followers than non-depressed. These features were then included in a classification model and achieved 82\% precision with 72\% accuracy. 

It is possible to identify depression and other mental health conditions through the analysis of Twitter posts. While \cite{coppersmith2014quantifying} provided a technique to gather open data, it did not detail a method to evaluate unlabelled population level depression. Whereas \cite{de2013predicting} used proprietary data, they developed a useful technique to extract features and analyse tweets to detect signals of mental health issues. This study will draw on the best-practices from the existing literature to classify a labelled dataset related to psychological distress. We then use transfer learning to estimate psychological distress status in unlabelled Brexit tweets.  

\subsection{Transfer Learning Approach for Sentiment Analysis}

Transfer learning is a machine learning technique which applies prevailing knowledge to decipher problems in other domains and subsequently present advanced prediction results \cite{liu2019survey}. This approach works for classification, regression, and clustering problems and can perform with no or limited labelled data \cite{daval2018epita}. In binary classification transfer learning, a model is trained on a source dataset ($D_s$) and then is used to classify a target dataset ($D_t$) even though $D_t$ comes from a different domain. For transfer learning to work however, $D_s$ and the $D_t$ must be related \cite{pan2010survey}. Transfer learning based-models are beneficial for cross-domain tasks particularly if there is poor data availability in a target domain as a high-performance model can be trained by applying readily available data from a source domain \cite{weiss2016survey}. 

Transfer learning has been especially successful in the field of sentiment analysis. For example, \cite{ bataa2019investigation} applied transfer learning to binary and multi-class sentiment classification on the Rakuten product review and Yahoo movie review datasets to reveal that these methods out-performed task-specific models. \cite{calais2011bias} applied transfer learning for \textit{opinion holder bias prediction} in an example of topic-based real time sentiment analysis. \cite{dong2018helping} utilised transfer learning to prompt sentiment embedding that portrayed sentiment polarity in multiple domains across diverse sentiment analysis tasks. Transfer learning has also been successful when applied to social media data, having been used to classify positive/negative/neutral sentiment, to detect adverse reactions, and to identify irony \cite{daval2018epita, alhuzali2019improving, zhangairony}. 

Transfer learning is beneficial when combined with rich representations of words and their perspectives, particularly when there is a large amount of vocabulary intersection between the two sets of data used in the experiment. In this paper, we used an Inductive Transfer Learning method as presented in \ref{sec:methods}.

\section{Proposed Method}
\label{sec:methods}
\subsection{Inductive Transfer Learning}

As described in \cite{pan2010survey} there are different techniques for transfer learning which fit the extent of class labelling available in source $D_s$ and target $D_t$ datasets. In \textit{Inductive Transfer learning}, labels of $D_s$ and $D_t$ are available, with source and target domains as the same. In \textit{Transductive Transfer Learning}, labels of $D_s$ are available however unavailable in $D_t$, with source and target domains as different but related. In \textit{Unsupervised Transfer Learning}, labels of $D_s$ and $D_t$ are unavailable, with source and target domains as different but related.

Our framework uses \textit{Inductive Transfer Learning} to improve the learning of the target predictive function $f_t(\cdot)$ in $D_t$ by using the knowledge in $D_s$ and $T_s$, where $T_s$ = $T_t$ (same classification task). 

\subsection{Source and Target Domains}
In this study, the source domain, $D_s$, represents individuals who had self-identified with anxiety or depression (collectively termed as `psychological distress'), with a learning task $T_s$. The target domain, $D_t$, represents a separate second set of tweets related to Brexit  with a learning task $T_t$

\subsubsection{{Source Domain ($D_s$)}}
Data for the source domain is collected for users who self-identified with depression or anxiety \cite{coppersmith2014quantifying}. Explicit keywords such as ``\textit{I was diagnosed with depression (anxiety)}'', ``\textit{I have depression (anxiety)}'' or \textit{``I’m depressed (anxious)}'' were selected to represent anxiety and depression at a non-clinical level of severity, but still indicative of a general level psychological distress. Here, we provide three example samples of tweets in the source domain representing the distress group (1) "\textit{Damaged people learn to adapt because we have to. We never fit in, never feel comfortable.}", (2) "\textit{I'm sad. I struggle daily with trying to find happiness. I will never be good enough}." (3) "\textit{You are not alone. I was diagnosed with anxiety too.}"

\subsubsection{Target Domain ($D_t$)}
The  goal of this study is to measure psychological distress as revealed in Twitter posts about Brexit. Therefore, tweets were collected which used \#Brexit. Here, we provide three example samples of Brexit related posts. (1) "\textit{I am really worried about the impact of no-deal \#Brexit on the NHS}", (2) "\textit{What are these people doing? Backstabbing themselves every day. I want to get off this planet. \#prorogue \#Brexit \#BorisJohnson}" and (3) "\textit{I am angry and outraged at the decision to suspend parliament to stop MPs from representing us. \#Brexit}".

\subsection{Feature Engineering}
After collecting the relevant data of $D_s$ and $D_t$, feature engineering techniques were applied to create a set of features suitable for detecting signals of psychological distress in Twitter posts. For this, we followed the approach outlined by \cite{de2013predicting}, in which the Linguistic Inquiry Word Count (LIWC) was used to extract psychologically meaningful information from social media contents to detect depression in populations. LIWC is a validated psycholinguistic tool, which processes a text and provides the percentage of words reflecting a range of social and psychological categories such as emotions, thinking styles, social concerns and parts of speech \cite{pennebaker2001linguistic}. A sample of the extracted features used in these experiments is included in Table \ref{features}. Note that, there are two general classes of features, the first describes qualities of the Twitter post and the second describes qualities of the Twitter user. 

\begin{table}[htp!]
\small
\caption{Post and User Features}
\label{features}
\begin{tabular}{|p{1.5cm}|p{6.5cm}|}
\hline
Emotion& User emotional state is extracted from the tweet using sentiment analysis. LIWC was used to extract Positive and Negative Affect, Anxiety, Anger and Sadness.                                                                                \\ \hline
Time& Time of the post is used to create an index of  whether the post occurred at night or day.  Following previous literature, night was defined as any post between 9PM and 5:59AM. The index was assigned for 1 night and -1 for day.           \\ \hline
Linguistic Style & Mental state has been shown to be predicted style of writing. LIWC was used to extract signals of linguistics style such as: articles, verbs, conjunctions, impersonal pronouns, personal pronouns, ,prepositions, certainty and quantifiers. \\ \hline
N-grams& Key words used in the posts were kept in the form joint unigrams present in source and target dataset \\ \hline
Engagement& The user's behaviour on twitter, encompassing the total volume of tweets by the user, the proportion which were retweets, and those which were replies.                                                                                        \\ \hline
Ego Network& the users' level of social connectedness was measured using their network. The number of followers of the user (inlinks) and the number followed by each user (outlinks).
\\ \hline
\end{tabular}
\end{table}

\paragraph{Feature Selection}
The extracted dataset included over 470 features, which is considerably big and might mislead the learning process. Hence, we used feature selection to find the most relevant features for the classification task. A feature was selected if it satisfied a set of criteria. The feature selection process produced a total of 234 features (from 470). The criteria of selected features are as follows: 
\begin{itemize}
\item Keep only those stemmed n-grams which covered 99\% of distress tweets.
\item Limit the source and training data to features which appear in both datasets. 
\item Eliminate any features which are highly correlated with each other (r2 > 0.75). 
\item Remove unimportant post level information such as date posted and user id. 
\end{itemize}

\subsection{Domain Transfer}
Domain transfer, also known as domain adaptation, is a technique in transfer learning in which the task remains the same but there is a domain shift or a distribution change between the source and the target. As an example, consider a model that has learned to classify reviews on electronic products for positive and negative sentiments, and is used for classifying the reviews for hotel rooms or movies. The task of sentiment analysis remains the same, but the domain (electronics and hotel rooms) has changed.

The intuitive approach to address domain transfer is to transform the source data so that the feature space between the source and target data matched. However, this could result in negative transfer between the source and target data. Negative transfer is when the target prediction is less than the no information prediction \cite{pan2010survey}. \cite{shimodaira2000improving} showed that the model will not predict optimally if the features of the two datasets have different marginal distributions. Two-sample Kolmogorov-Smirnoff tests were conducted to test for differences between the source and target data, and results for all features were significant at p < 0.001. These results showed that we reject the null hypothesis that the samples were drawn from populations with the same distributions. As such, the source and target data were shown to come from different domains and additional steps were needed to minimise differences between source and target data.

There are numerous approaches to minimising differences between source and target data when they come from significantly different domains \cite{borgwardt2006integrating}. There is a field of study dedicated to helping alleviate such problems of domain adaptation \cite{ben2010theory}. In our proposed method, three steps were taken to minimise these differences. These steps are stated in Algorithm \ref{alg}, step 2. 

\begin{itemize}
  \item Re-sample source data so that the conditional probability of distress and standard posts match those of the target data. 
   \item Remove user features from source and target data as the differences in methodology between the collection of source and target data resulted in systematic differences between the distributions of followers, followees, total tweets and total favourites.
    \item Re-weight source data using the mean difference weight between source and target data so that the mean for all features of the weighted source data equals those of the target data.
\end{itemize}	
 
\subsection{Supervised Learning Models}
\label{class}
The prepared psychological distress (source) data was used to train a supervised machine learning algorithm classifier which identified psychological distress signals in Brexit tweets. Multiple classifiers were attempted and the best performing classifier was selected to predict distress labels for the full set of unlabelled Brexit (target) data. 

\begin{figure}[htp!]
\begin{center}
\includegraphics[width = \columnwidth]{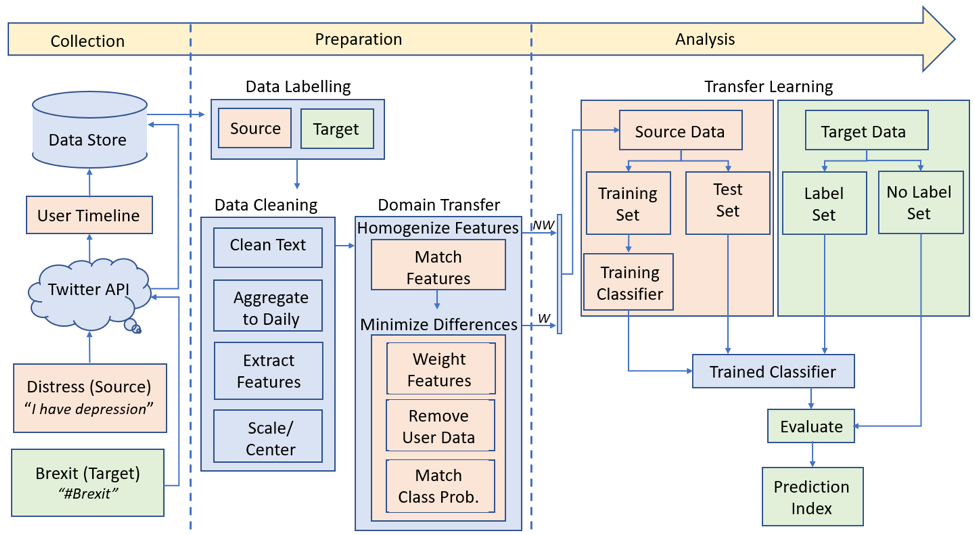}
\caption{Architectural Framework of Methodology}
\label{archi}
\end{center}
\end{figure}

The overall architecture of the proposed method is illustrated in Figure \ref{archi} for weighted (W) and unweighted (NW) source data conditions. 

\subsection{Brexit Distress Index}
To monitor changes in psychological distress over time, we compared results for the target time period using an index of predicted values. Indexes have been used in similar research to compare results of predictive models on population level trends \cite{de2013predicting, bagroy2017social}. Brexit Distress Index (BDI) is defined as the standardised difference between the number of distress-indicative posts and ``standard'' posts across an observed time period.  
Brexit Distress Index is calculated as follows: 
\begin{equation}
BDI=\frac{n_d (t)-\mu_d}{\alpha_d} -  \frac{n_s (t)-\mu_s}{\alpha_s} 
\end{equation}
where $n_d (t)$ and $n_s (t)$ are the number of distress-indicative and ``standard'' posts, while $\mu_d (\mu_s)$ and $\alpha_d (\alpha_s)$ are the mean and standard deviations of the number of corresponding posts shared in a fixed time period.

\subsection{Algorithm Description}
Algorithm \ref{alg} describes the steps of the proposed method. The first step of the algorithm combines features from $D_s$ (labelled distress-indicative data) and $D_t$ (partially labelled Brexit data). When all features present in the source data are also available in the target data, it is called Homogeneous Transfer Learning \cite{weiss2016survey}. 
Domain adaptation is presented in step 2. 

The labelled psychological distress dataset was used to train a distress classification model, which was then used to improve the classification of distress in Brexit tweets. In addition, a small sample of labelled Brexit tweets provide a level of accuracy for the inference to the full set of unlabelled Brexit tweets. 

\begin{algorithm}
\caption{Inductive Transfer Learning with Brexit Distress Index}\label{euclid}
\begin{itemize}
    \item \textbf{Input 1:} Source dataset $D_s^n$ with labels $\{(x_1,y_1), \dots, (x_n,y_n)\}$. $x_i$ is the $i_{th}$ Twitter post for self-identified distress users (or control) and $y_i$ is a distress label for the post.
    \item  \textbf{Input 2:} Target dataset $D_t^m$ with no labels $\{(x_1,b_1)\dots, (x_m,b_m)\}$. Here, $x_i$ is the $i_{th}$ Twitter post using $\#$Brexit and $b_i$ is the yet to be predicted distress label for the post.
     \item \textbf{Output:} Brexit Distress Index $BDI$ for a day $t$.
\end{itemize}
\begin{algorithmic}[1]
\item Select common features $f$ from $D_s^n$ and $D_t^m$

\item Domain Adaptation: to minimise marginal and conditional probability distribution differences between $D_s^n$ and $D_t^m$.
\begin{algorithmic}[1]
    \item  Re-sample $D_s^n$ to minimise the difference in conditional probability between $D_s$ and  $D_t$.
        \item  Scale and centre all features $\in f$ for $D_s^n$ and $D_t^m$.
    \item Re-weight features $f$ in $D_s^n$ to minimise the difference of mean between $D_s^n$  and  $D_t^m$.
\end{algorithmic} 
\item Split $D_s^n$  into training and test datasets, $D_s^{train}$ and $D_s^{test}$
\item Train classifiers $C_1 \dots, C_j$ by the source dataset  $D_s^{train}$. Choose optimal parameters to predict $y_{test}$ for $D_s^{test}$ to maximise accuracy. 
\item Manually label a sample of $D_t^m$ of size k where k<< m
\item $\forall x \in D_t^k$ predict target labels $(b^\prime=b^{\prime}_1 \dots ,b^{\prime}_k)$ using $C_1 \dots ,C_j$. 
\item Select the best classifier from $C_1 \dots,C_j$ which maximise the sum of Sensitivity and Specificity on $D_t^k$ 
\item $\forall x \in D_t^m$ predict target labels
$(b^{\prime}=b^{\prime}_1 \dots ,b^{\prime}_m)$ using the selected classifier.
\item Calculate Brexit Distress Index for $D_t^m$ 

\end{algorithmic}
\label{alg}
\end{algorithm}

\section{Experiments and results}

\subsection{Datasets}
\subsubsection{{Psychological distress dataset (source domain):}} An effective technique for gathering samples of labelled mental health data from social media was described in \cite{coppersmith2014quantifying}. To create the labelled distress group, the Twitter API was queried for users who self-identified with depression or anxiety using key phrases such as “I was diagnosed with depression/anxiety”, “I have depression/anxiety” or “I’m depressed/anxious”. The search returned A total of {10,000,000} tweets which were then manually validated to remove disingenuous self-identification statements (such as those interpreted to represent irony or sarcasm rather than genuine distress). The cleansing process resulted in a sample of {868} users, who showed positive indicators of psychological distress.

The Twitter API was then used to collect the tweet history of each identified user. The Twitter API allows access to the most recent 3,000 tweets for each user. The full history of tweets for each user was then limited to the most recent 3-months of posts.
Considering the Twitter privacy settings of each user, and keeping only users with 25 or more tweets, the  dataset contained tweets for 639 users posting 735,599 tweets which were labelled as Positive (distress-indicative). 

In addition, a random sample of similarly sized Twitter users (649 users) were used to collect the Tweet history, posting during the same time period, and ensuring that they had no psychological distress signals. This process yielded a control group dataset for 649 users and 735,599 tweets which were labelled as negative (non-distress-indicative). The complete psychological distress dataset (source data) contained historical tweets for over 1,200 users labelled as positive or negative. The source domain dataset is described in Table \ref{distable}. 


\subsubsection{Brexit dataset (target domain):}
Tweets were collected which used \#Brexit. Following Twitter’s Terms of Service, only three weeks of recent tweets are available for free using the Twitter API \cite{hernandez2018api}. The Brexit dataset contained a large corpus of over 260,000 tweets for 83,489 users. In contrast with the above psychological distress dataset, it was not possible to manually label all Brexit tweets as non-/distress-indicative due to the lack of clear indicative contents. However, a smaller subset of 300 tweets was manually labelled as being distress-indicative (positive) or  negative (non-distress-indicative). These posts were distinct in that they included explicit sentences describing users depressive status which helped us to assign labels. This smaller subset was used as a proxy for transfer learning accuracy. 

The collected Brexit dataset included three consecutive periods between August 2019 and September 2019 with \#Brexit. We particularly focused on this time period because it was just before the initial date of EU exit (October 31st). At this time there was an increased wave of fear and uncertainty among UK residence around No-Deal Brexit. 



As noticed, the number of users in the Brexit dataset was significantly larger than those in the Distress dataset. In contrast, the number of posts in the Brexit dataset were less than the ones in the Distress dataset. On average, a user had 1,153 posts in the Distress dataset, and only 3.22 posts in the Brexit data. This dataset is described in Table \ref{distable}. 


\begin{table}[ht!]
\caption{Datasets description}
\label{distable}
\begin{tabular}{|p{3.2cm}|p{2.2cm}|p{2cm}|}
\hline& Psychological distress dataset (source domain) & Brexit dataset (Target domain) \\ 
\hline
Total number of users & 1,279& 83,489\\ 
 \# users with distress &   639 &   103* \\ 
 \# users in control group &   649&   197*  \\ 
 \hline
Total number of tweets& 1,475.95 & 269,134 \\ 
 Positive (distress)&  735,599 &  103*\\ 
 Negative (no-distress)&  740,348 &  197* \\ \hline
Mean posts/ user& 1,153& 3.22 \\ 
Mean posts/day/user& 26.94& 1.18\\ 
\hline
\end{tabular}

\end{table}

\subsubsection{Data Preparation:} 
Relevant Twitter API data related to user and post meta-data were selected from source and target domains. Text were cleaned and transformed into a set of reproducible features, as described in section \ref{sec:methods}. This resulted in two different source datasets, which were compared in the analysis: the first included matched features between the source and target data (denoted here as unweighted dataset), and the second contained matched features after applying the domain adaptation technique, which minimised the probability distribution differences between the source and target data, (denoted here as weighted dataset).  

The collected dataset included the timestamp of the tweet, language, user information such as friends, followers, and favourites, profile time zone, time account created, and others explained below. In addition, all screen names and user identifiers were anonymized prior to analysis. Complying with GDPR standards, the anonymization technique utilised salted hashing \cite{Burgess2018gdpr}.
The raw corpus of twitter posts was cleaned by filtering the tweets to only include those written in English, converting to lowercase, removing all numbers, stop words, punctuation, whitespace, URLs and other Twitter API codes (@, amp), and finally stemming words. For numeric features, data were scaled and centred. 
Data were extracted from the API at the post-level. Thus, each row in the dataset represented a unique post from a user. Prior to analysis, data were aggregated to the daily level such that each row in the dataset contained concatenated text for the full day of twitter activity in addition to corresponding aggregated meta-data for each individual tweet. 

\subsection{Performance Evaluation}
The aim of this experiment was to evaluate the effectiveness of the transfer learning approach as well as the domain adaptation in detecting psychological distress signals in Brexit tweets using supervised classification techniques. The predictive ability of the extracted features were assessed using three traditional classifier algorithms (SVM, LR, and RF), as discussed in section \ref{class}. 

Support Vector Machines (SVM) use a set of flexible parameters to create an optimal hyperplane between classes to achieve accurate results. The SVM algorithm has been used in detecting mental health signals, as well as in transfer learning \cite{bagroy2017social, bahadori2011learning}. However, SVMs are sensitive to input features as well as to changes in parameters \cite{lantz2013machine}. Accurate SVMs require proper parameter settings. This research uses an SVM algorithm with a radial bases function (RBF) kernal. The two main parameters of this pairing are ‘C’ to assign a penalty for error in the SVM algorithm, and ‘sigma’ to determine the effect of error in the RBF function. A grid search technique was used on 'C' between 0.25 and 64, and 'sigma' between 0.001 and 0.5 to improve classifier accuracy.

Logistic Regression (LR) is an application of the regression equation suited for predicting the probability of a binary outcome. This model specifies the relationships between the binary dependent variable, and the predictor variables \cite{lantz2013machine}. The result provides the probability of the outcome along with significance of each predictor.

Random Forests (RF) are an ensemble-based method which combine the results of N number of decision trees to create a stronger learner. The model is comprised of multiple decision trees, each made up of random subsets of features, which are used to vote and combine the trees’ predictions \cite{lantz2013machine}. Due to trees being comprised of random subsets of variables, this method is robust to high dimensional feature spaces. 

Classifiers performance was assessed in terms of accuracy ({TP + TN}/{TP + TN + FP + FN}), sensitivity ({TP}/{TP + FN}) and specificity ({TN}/{TN + FP}),
where TP, FP , TN and FN denotes true positives, false positives, true negatives and false negatives, respectively. 
A grid-search technique was applied to find the optimal hyperparameters in all the studied learning algorithms. The experiments were completed using an Intel Core i7-8705G CPU machine with eight cores in parallel.

Our results from the transfer learning approach are reported in two sections. The first section reports the predictive performance of the unweighted source data without applying the domain adaptation, whereas the second section reports results of the weighted source data after applying the domain adaptation. The tests results of all learning models were the result of 5-Fold Cross Validation. 
Source and target data were scaled and centered transformations of raw data. In addition, distress and control classes were even, making a balanced classification problem.
In addition, model transfer accuracy is compared between two conditions of source data (weighted, unweighted). As described in \ref{sec:methods}, in the weighted condition domain adaptation techniques were used to minimize the difference between source and target prior to model training. This resulted in an unbalanced classification problem with the ratio of Control to Distress class labels being 3:1. The unweighted condition does not use domain adaptation, in addition, the ratio of Control to Distress class labels is 1:1. 

As seen in Table \ref{resultsTab1}, results of the unweighted condition showed moderate accuracy for both SVM and LR models on the source data. However, the RF model included all features which were included in the previous models, and significantly outperformed the other methods on source data. When applied to the sample of Brexit tweets (target), all models lost substantial accuracy. This was especially true for RF. Loss of accuracy during transfer learning, called negative transfer, was addressed in the weighted source condition. 

In the weighted condition, Table \ref{resultsTab2}, loss in accuracy for transfer learning to Brexit dataset was substantially less than those of the unweighted source data. For example, LR and SVM achieved similar accuracy on the unseen target data as they did on the source testing data. Negative transfer still occurred for RF.


\begin{table}[!htp]
\footnotesize
  \begin{center}
    \caption{Results from transfer learning (unweighted source data without applying the domain adaptation)}
    \label{resultsTab1}
    \begin{tabular}{|l|c|c|c|c|c|}
    \hline
      Model & Data & accuracy & Specificity & Sensitivity  \\
      \hline
      \multirow{2}{4em}{SVM} & Source & 0.65 & 0.47 & 0.81  \\ 
      & Target & 0.43 & 0.22 & 0.86 \\ 
      \hline
      \multirow{2}{4em}{LR}& Source & 0.62 & 0.52 & 0.70 \\ 
      & Target & 0.46 & 0.48 & 0.41  \\ 
      \hline
      \multirow{2}{4em}{RF} & Source & 0.96 & 0.97 & 0.96  \\ 
      & Target & 0.44 & 0.46 & 0.42  \\
      \hline
    \end{tabular}
  \end{center}
\end{table}

\begin{table}[!htp]
\footnotesize
  \begin{center}
     \caption{Results from transfer learning (weighted source data after applying the domain adaptation)}
    \label{resultsTab2}
    \begin{tabular}{|l|c|c|c|c|c|}
    \hline
      Model & Data & accuracy & Specificity & Sensitivity  \\
      \hline
      \multirow{2}{4em}{SVM} & Source & 0.66 & 0.94 & 0.13 \\ 
      & Target & 0.66 & 1 & 0\\ \hline
      \multirow{2}{4em}{LR}& Source & 0.66 & 0.91 & 0.20 \\ 
      & Target  & 0.62 & 0.73 & 0.41 \\  \hline
      \multirow{2}{4em}{RF} & Source  & 0.67 & .90 & 0.23 \\ 
      & Target & 0.45 & 0.63 & 0.10 \\ \hline
    \end{tabular}
  \end{center}
\end{table}

Weighted and unweighted conditions achieved similar accuracy when classifying source data. However, the most effective model for transfer learning, as indicated by accuracy (62\%), specificity (73\%) and sensitivity (41\%) for classifying Brexit tweets, was LR with weighted source data. This model will be used in the next section to identify rates of psychological distress in unlabelled Brexit related tweets. 



\begin{figure}[htb]
\vskip 0.2in
\begin{center}
\centerline{\includegraphics[width=\columnwidth, height=6.5cm]{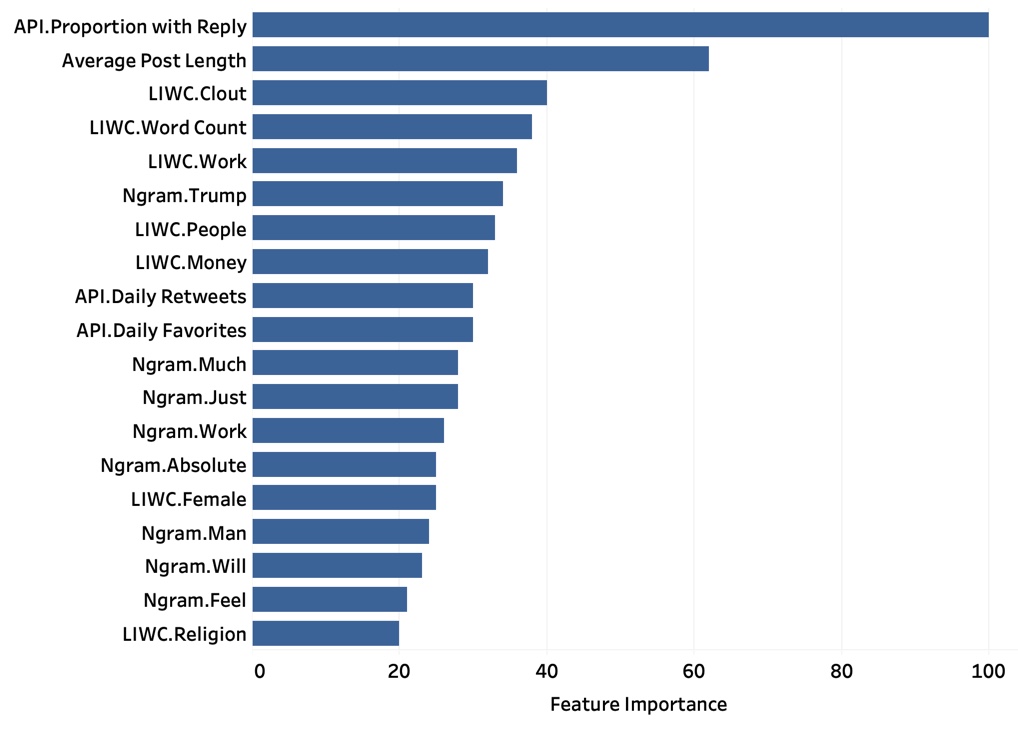}}
\caption{Top features for most accurate model with weighted source data.}
\label{fig:mapping}
\end{center}
\vskip -0.2in
\end{figure}

We also applied a feature selection method to identify the most relevant features using the best performing classifier (Logistic Regression). Figure \ref{fig:mapping} ranks the predictive features according to their weight importance. The analysis reveals that the following features act as strong predictors in detecting the distress status in the Brexit dataset: (1) proportion of posts with reply, (2) average length of posts, (3) LIWC Clout \footnote{Clout is defined as confidence and social status displayed in writing or talking.} and (4) the LIWC Word Count. 

\subsection{Brexit Distress Index}
\begin{figure}[htb]
\vskip 0.2in
\begin{center}
\centerline{\includegraphics[width=\columnwidth]{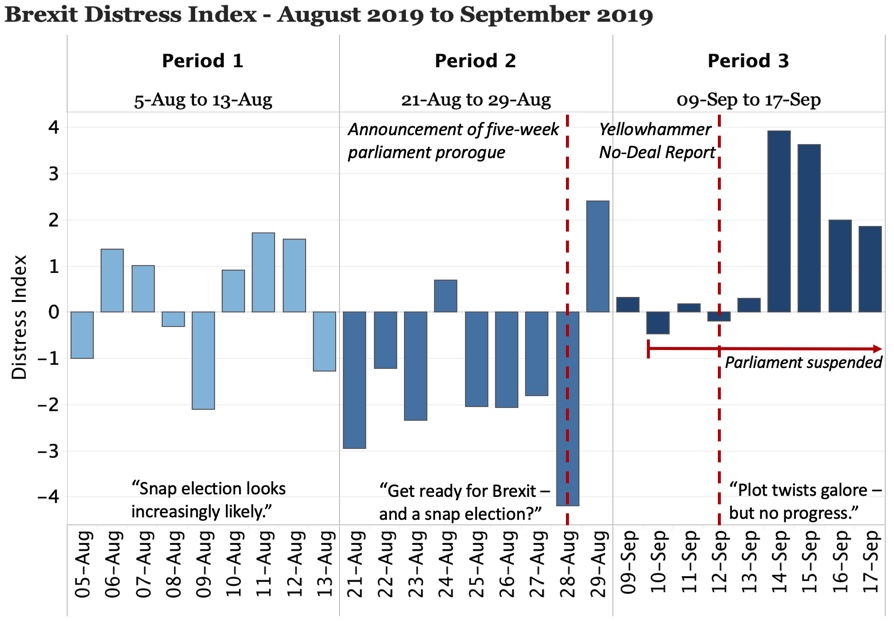}}
\caption{Change in Brexit Distress from August 2019 to September 2019.}
\label{fig:mapping2}
\end{center}
\vskip -0.2in
\end{figure}

Using a Logistic Regression Model trained on weighted source data which was applied applied to classify unlabelled Brexit tweets, an index was created to visualize change in psychological distress related to Brexit over time. Figure \ref{fig:mapping2} shows the results of transfer learning distress labels applied to unlabelled \#Brexit data and how these trends changed over time. Positive index values indicate times of more distressed tweets than average for the time period. Change in the index is plotted against major events in \#Brexit news, made evident by major events (red lines) with corresponding weekly Brexit headlines from The Guardian (black quotations). 
Keeping in mind the error of the transfer learning model, the trends in psychological distress correspond to major Brexit news headlines. For example, the overall trend of increased distress over time correlates with the announcement of news that Parliament would be prorogued for the longer than usual period of five weeks \cite{Lyons2019pro, bbc2019pro}. Distress levels remained stable through September until the announcement, and then increased following the announcement. In addition, the release of ``Operation Yellowhammer'', a government report predicting possible negative consequences of No-Deal Brexit was shortly followed by another increase in distress which remained stable until the end of the observed period \cite{Middleton2019yellow}. Trends correspond to Guardian headlines for Brexit news in that distress increased with increased uncertainty of news headlines. Overall, uncertainty expressed in the language of headlines, as well as distress signals increased from Periods 1 to 3.

\section{Discussion}
\label{sec:discussion}
Major political fluctuations may have temporary or longer-lasting effects on the psychological well-being of individuals.
The goal of this research was to detect non-clinical psychological distress signals in the UK using Brexit Twitter data and to compare these results to major Brexit events. However, obtaining ``Gold Standard'' training data for people with psychological distress (depression/anxiety) is a challenging task. 

To address this, we used an open source approach to collect distress-indicative tweets which relied upon users who had self-identified as having depression or anxiety. A supervised machine learning model was trained on such data and applied to unseen Brexit data using a transfer learning technique. The extent to which distress would be successfully identified in Brexit tweets relied upon two important steps: the first was to accurately classify distress tweets from standard tweets, and the second was to accurately transfer knowledge from available labelled distress data to unlabelled Brexit data. These experiments applied a transfer learning approach using traditional machine learning classifiers to the problem of detecting distress signals in social media. There are also promising transfer learning approaches based on deep learning techniques which may be well-suited to this problem, and these should be considered in future research \cite{alhuzali2019improving,bataa2019investigation, daval2018epita}. 

This research extended upon the work from \cite{de2013predicting} and \cite{coppersmith2014quantifying, coppersmith2015adhd, coppersmith2015clpsych} by applying results to identifying signals of psychological distress for tweets related to a specific hashtag. 
When it was discovered that the distress and Brexit datasets were too dissimilar to achieve accurate transfer learning results, a domain adaptation technique was utilised to minimise these differences. This process resulted in a model which was comparably accurate to the previous model (67\% Accuracy, 91\% Sensitivity, 20\% Specificity) in classifying labelled distress data, but much more effective in generalising previously unseen Brexit data (62\% Accurate, 73\% Sensitivity, 41\% Specificity). 

The next step in the experiment was to transfer the trained model to a set of unlabelled data. When the transfer learning model was applied to a large set of unlabelled recent Brexit tweets, trends compared favourably to major Brexit headlines. Considering the model is moderately accurate in classifying labelled data, results appear somewhat effective in identifying population level psychological distress signals in Brexit tweets. Other sentiment analysis techniques show spikes in activity related to major political, cultural, economic, and natural events \cite{bollen2011modeling, kim2009detecting, lampos2013analysing, nguyen2013royal}. Major events during the observation period, such as the surprise announcement of an uncharacteristic five-week parliament suspension, and the release of official predictions of worst-case scenario effects of No-Deal Brexit, correlated with increased psychological distress. One note on this relationship is that there was a time lag between the date of event and the change in public sentiment. An explanation for this is that it may take time for news of the event to reach the general public. For example, although the parliament prorogation was approved on August $28^{\text{th}}$ 2019, it was not until August $29^{\text{th}}$ that the event was on the front page of British papers \cite{lyons2019prorogue}. 
In addition to spikes in distress around major events, these results show that there was an increase in general levels of distress related to Brexit from the beginning of August until mid-September. 
At that time there was increased uncertainty around how the UK would part with the EU and whether this would result in a favourable deal. This research shows that such uncertainty corresponds to a measurable negative impact on the psychological well-being of UK residents. 

Although Brexit is a strong political issue, the results from this analysis reveal that residents across the political spectrum express distress related to the event. As made evident by the tweets in this sample, some expressed fear and sadness related to leaving the EU and possible downfalls of the event. Whereas others expressed anger that it has taken so long for the government to follow through with the result of the Brexit referendum. The results provide quantified evidence of the possible negative psychological effects of a No-Deal Brexit. The polarising effects of such an event may lead to additional negative effects for those living in the UK regardless of their political position because, as made evident by the results of the current study, this has become an increasingly personal issue.

\section{Conclusion}
\label{sec:conclusions}

Some political events can have temporary or longer lasting effects on the mental well-being of individuals. This paper proposes a sentiment analysis approach based on transfer learning to measure non-clinical psychological distress in Brexit tweets. The transfer learning was able to apply knowledge already learned in the source domain (tweets for users who self-identified with depression or anxiety) to the target domain (tweets about Brexit). A domain adaptation technique was used to decrease the impact of negative transfer between source and target domains. The proposed Brexit Distress Index was able to identify changes in distress that correlated with major events such as the announcement of prorogation in August 2019. 

As much of the public discourse about Brexit is focused on possible economic impacts, this research gives quantifiable evidence for a discourse which should also include concerns for the psychological well-being of people living in the UK during this historic time of political change.
The significant benefit of the present research is that it ventured into a repeatable approach to analysis in which a classifier is trained to identify users with psychological distress and is then applied to find signs of distress in other domains.


\bibliographystyle{ACM-Reference-Format}
\bibliography{references.bib} 


\begin{thebibliography}{40}


\ifx \showCODEN    \undefined \def \showCODEN     #1{\unskip}     \fi
\ifx \showDOI      \undefined \def \showDOI       #1{#1}\fi
\ifx \showISBNx    \undefined \def \showISBNx     #1{\unskip}     \fi
\ifx \showISBNxiii \undefined \def \showISBNxiii  #1{\unskip}     \fi
\ifx \showISSN     \undefined \def \showISSN      #1{\unskip}     \fi
\ifx \showLCCN     \undefined \def \showLCCN      #1{\unskip}     \fi
\ifx \shownote     \undefined \def \shownote      #1{#1}          \fi
\ifx \showarticletitle \undefined \def \showarticletitle #1{#1}   \fi
\ifx \showURL      \undefined \def \showURL       {\relax}        \fi
\providecommand\bibfield[2]{#2}
\providecommand\bibinfo[2]{#2}
\providecommand\natexlab[1]{#1}
\providecommand\showeprint[2][]{arXiv:#2}

\bibitem[\protect\citeauthoryear{Alhuzali and Ananiadou}{Alhuzali and
  Ananiadou}{2019}]%
        {alhuzali2019improving}
\bibfield{author}{\bibinfo{person}{Hassan Alhuzali} {and}
  \bibinfo{person}{Sophia Ananiadou}.} \bibinfo{year}{2019}\natexlab{}.
\newblock \showarticletitle{Improving classification of adverse drug reactions
  through using sentiment analysis and transfer learning}. In
  \bibinfo{booktitle}{\emph{Proceedings of the 18th BioNLP Workshop and Shared
  Task}}. \bibinfo{pages}{339--347}.
\newblock


\bibitem[\protect\citeauthoryear{Bagroy, Kumaraguru, and De~Choudhury}{Bagroy
  et~al\mbox{.}}{2017}]%
        {bagroy2017social}
\bibfield{author}{\bibinfo{person}{Shrey Bagroy}, \bibinfo{person}{Ponnurangam
  Kumaraguru}, {and} \bibinfo{person}{Munmun De~Choudhury}.}
  \bibinfo{year}{2017}\natexlab{}.
\newblock \showarticletitle{A social media based index of mental well-being in
  college campuses}. In \bibinfo{booktitle}{\emph{Proceedings of the 2017 CHI
  Conference on Human factors in Computing Systems}}.
  \bibinfo{pages}{1634--1646}.
\newblock


\bibitem[\protect\citeauthoryear{Bahadori, Liu, and Zhang}{Bahadori
  et~al\mbox{.}}{2011}]%
        {bahadori2011learning}
\bibfield{author}{\bibinfo{person}{Mohammad~Taha Bahadori},
  \bibinfo{person}{Yan Liu}, {and} \bibinfo{person}{Dan Zhang}.}
  \bibinfo{year}{2011}\natexlab{}.
\newblock \showarticletitle{Learning with minimum supervision: A general
  framework for transductive transfer learning}. In
  \bibinfo{booktitle}{\emph{2011 IEEE 11th International Conference on Data
  Mining}}. IEEE, \bibinfo{pages}{61--70}.
\newblock


\bibitem[\protect\citeauthoryear{Bataa and Wu}{Bataa and Wu}{2019}]%
        {bataa2019investigation}
\bibfield{author}{\bibinfo{person}{Enkhbold Bataa} {and}
  \bibinfo{person}{Joshua Wu}.} \bibinfo{year}{2019}\natexlab{}.
\newblock \showarticletitle{An investigation of transfer learning-based
  sentiment analysis in japanese}.
\newblock \bibinfo{journal}{\emph{arXiv preprint arXiv:1905.09642}}
  (\bibinfo{year}{2019}).
\newblock


\bibitem[\protect\citeauthoryear{BBC}{BBC}{2019}]%
        {bbc2019pro}
\bibfield{author}{\bibinfo{person}{BBC}.} \bibinfo{year}{2019}\natexlab{}.
\newblock \showarticletitle{Prorogation: How can the government suspend
  Parliament?}
\newblock
  \bibinfo{journal}{\emph{https://www.bbc.co.uk/news/uk-politics-48936711}}
  (\bibinfo{year}{2019}).
\newblock


\bibitem[\protect\citeauthoryear{Ben-David, Blitzer, Crammer, Kulesza, Pereira,
  and Vaughan}{Ben-David et~al\mbox{.}}{2010}]%
        {ben2010theory}
\bibfield{author}{\bibinfo{person}{Shai Ben-David}, \bibinfo{person}{John
  Blitzer}, \bibinfo{person}{Koby Crammer}, \bibinfo{person}{Alex Kulesza},
  \bibinfo{person}{Fernando Pereira}, {and} \bibinfo{person}{Jennifer~Wortman
  Vaughan}.} \bibinfo{year}{2010}\natexlab{}.
\newblock \showarticletitle{A theory of learning from different domains}.
\newblock \bibinfo{journal}{\emph{Machine learning}} \bibinfo{volume}{79},
  \bibinfo{number}{1-2} (\bibinfo{year}{2010}), \bibinfo{pages}{151--175}.
\newblock


\bibitem[\protect\citeauthoryear{Benton, Mitchell, and Hovy}{Benton
  et~al\mbox{.}}{2017}]%
        {benton2017multi}
\bibfield{author}{\bibinfo{person}{Adrian Benton}, \bibinfo{person}{Margaret
  Mitchell}, {and} \bibinfo{person}{Dirk Hovy}.}
  \bibinfo{year}{2017}\natexlab{}.
\newblock \showarticletitle{Multi-task learning for mental health using social
  media text}.
\newblock \bibinfo{journal}{\emph{arXiv preprint arXiv:1712.03538}}
  (\bibinfo{year}{2017}).
\newblock


\bibitem[\protect\citeauthoryear{Bollen, Mao, and Pepe}{Bollen
  et~al\mbox{.}}{2011}]%
        {bollen2011modeling}
\bibfield{author}{\bibinfo{person}{Johan Bollen}, \bibinfo{person}{Huina Mao},
  {and} \bibinfo{person}{Alberto Pepe}.} \bibinfo{year}{2011}\natexlab{}.
\newblock \showarticletitle{Modeling public mood and emotion: Twitter sentiment
  and socio-economic phenomena}. In \bibinfo{booktitle}{\emph{Fifth
  International AAAI Conference on Weblogs and Social Media}}.
\newblock


\bibitem[\protect\citeauthoryear{Borgwardt, Gretton, Rasch, Kriegel,
  Sch{\"o}lkopf, and Smola}{Borgwardt et~al\mbox{.}}{2006}]%
        {borgwardt2006integrating}
\bibfield{author}{\bibinfo{person}{Karsten~M Borgwardt},
  \bibinfo{person}{Arthur Gretton}, \bibinfo{person}{Malte~J Rasch},
  \bibinfo{person}{Hans-Peter Kriegel}, \bibinfo{person}{Bernhard
  Sch{\"o}lkopf}, {and} \bibinfo{person}{Alex~J Smola}.}
  \bibinfo{year}{2006}\natexlab{}.
\newblock \showarticletitle{Integrating structured biological data by kernel
  maximum mean discrepancy}.
\newblock \bibinfo{journal}{\emph{Bioinformatics}} \bibinfo{volume}{22},
  \bibinfo{number}{14} (\bibinfo{year}{2006}), \bibinfo{pages}{e49--e57}.
\newblock


\bibitem[\protect\citeauthoryear{Burgess}{Burgess}{2018}]%
        {Burgess2018gdpr}
\bibfield{author}{\bibinfo{person}{Matt Burgess}.}
  \bibinfo{year}{2018}\natexlab{}.
\newblock \showarticletitle{What is GDPR? The summary guide to GDPR compliance
  in the UK}.
\newblock
  \bibinfo{journal}{\emph{http://www.wired.co.uk/article/what-is-gdpr-uk-eu-legislation-compliance-summary-fines-2018}}
  (\bibinfo{year}{2018}).
\newblock


\bibitem[\protect\citeauthoryear{Calais~Guerra, Veloso, Meira~Jr, and
  Almeida}{Calais~Guerra et~al\mbox{.}}{2011}]%
        {calais2011bias}
\bibfield{author}{\bibinfo{person}{Pedro~Henrique Calais~Guerra},
  \bibinfo{person}{Adriano Veloso}, \bibinfo{person}{Wagner Meira~Jr}, {and}
  \bibinfo{person}{Virg{\'\i}lio Almeida}.} \bibinfo{year}{2011}\natexlab{}.
\newblock \showarticletitle{From bias to opinion: a transfer-learning approach
  to real-time sentiment analysis}. In \bibinfo{booktitle}{\emph{Proceedings of
  the 17th ACM SIGKDD international conference on Knowledge discovery and data
  mining}}. \bibinfo{pages}{150--158}.
\newblock


\bibitem[\protect\citeauthoryear{Coppersmith, Dredze, and Harman}{Coppersmith
  et~al\mbox{.}}{2014}]%
        {coppersmith2014quantifying}
\bibfield{author}{\bibinfo{person}{Glen Coppersmith}, \bibinfo{person}{Mark
  Dredze}, {and} \bibinfo{person}{Craig Harman}.}
  \bibinfo{year}{2014}\natexlab{}.
\newblock \showarticletitle{Quantifying mental health signals in Twitter}. In
  \bibinfo{booktitle}{\emph{Proceedings of the workshop on computational
  linguistics and clinical psychology: From linguistic signal to clinical
  reality}}. \bibinfo{pages}{51--60}.
\newblock


\bibitem[\protect\citeauthoryear{Coppersmith, Dredze, Harman, and
  Hollingshead}{Coppersmith et~al\mbox{.}}{2015a}]%
        {coppersmith2015adhd}
\bibfield{author}{\bibinfo{person}{Glen Coppersmith}, \bibinfo{person}{Mark
  Dredze}, \bibinfo{person}{Craig Harman}, {and} \bibinfo{person}{Kristy
  Hollingshead}.} \bibinfo{year}{2015}\natexlab{a}.
\newblock \showarticletitle{From ADHD to SAD: Analyzing the language of mental
  health on Twitter through self-reported diagnoses}. In
  \bibinfo{booktitle}{\emph{Proceedings of the 2nd Workshop on Computational
  Linguistics and Clinical Psychology: From Linguistic Signal to Clinical
  Reality}}. \bibinfo{pages}{1--10}.
\newblock


\bibitem[\protect\citeauthoryear{Coppersmith, Dredze, Harman, Hollingshead, and
  Mitchell}{Coppersmith et~al\mbox{.}}{2015b}]%
        {coppersmith2015clpsych}
\bibfield{author}{\bibinfo{person}{Glen Coppersmith}, \bibinfo{person}{Mark
  Dredze}, \bibinfo{person}{Craig Harman}, \bibinfo{person}{Kristy
  Hollingshead}, {and} \bibinfo{person}{Margaret Mitchell}.}
  \bibinfo{year}{2015}\natexlab{b}.
\newblock \showarticletitle{CLPsych 2015 shared task: Depression and PTSD on
  Twitter}. In \bibinfo{booktitle}{\emph{Proceedings of the 2nd Workshop on
  Computational Linguistics and Clinical Psychology: From Linguistic Signal to
  Clinical Reality}}. \bibinfo{pages}{31--39}.
\newblock


\bibitem[\protect\citeauthoryear{Daval-Frerot, Bouchekif, and
  Moreau}{Daval-Frerot et~al\mbox{.}}{2018}]%
        {daval2018epita}
\bibfield{author}{\bibinfo{person}{Guillaume Daval-Frerot},
  \bibinfo{person}{Abdessalam Bouchekif}, {and} \bibinfo{person}{Anatole
  Moreau}.} \bibinfo{year}{2018}\natexlab{}.
\newblock \showarticletitle{Epita at SemEval-2018 Task 1: Sentiment analysis
  using transfer learning approach}. In \bibinfo{booktitle}{\emph{Proceedings
  of The 12th International Workshop on Semantic Evaluation}}.
  \bibinfo{pages}{151--155}.
\newblock


\bibitem[\protect\citeauthoryear{De~Choudhury, Gamon, Counts, and
  Horvitz}{De~Choudhury et~al\mbox{.}}{2013}]%
        {de2013predicting}
\bibfield{author}{\bibinfo{person}{Munmun De~Choudhury},
  \bibinfo{person}{Michael Gamon}, \bibinfo{person}{Scott Counts}, {and}
  \bibinfo{person}{Eric Horvitz}.} \bibinfo{year}{2013}\natexlab{}.
\newblock \showarticletitle{Predicting depression via social media}. In
  \bibinfo{booktitle}{\emph{Seventh international AAAI conference on weblogs
  and social media}}.
\newblock


\bibitem[\protect\citeauthoryear{Dong and De~Melo}{Dong and De~Melo}{2018}]%
        {dong2018helping}
\bibfield{author}{\bibinfo{person}{Xin~Luna Dong} {and} \bibinfo{person}{Gerard
  De~Melo}.} \bibinfo{year}{2018}\natexlab{}.
\newblock \showarticletitle{A helping hand: Transfer learning for deep
  sentiment analysis}. In \bibinfo{booktitle}{\emph{Proceedings of the 56th
  Annual Meeting of the Association for Computational Linguistics (Volume 1:
  Long Papers)}}. \bibinfo{pages}{2524--2534}.
\newblock


\bibitem[\protect\citeauthoryear{Guma and Dafydd~Jones}{Guma and
  Dafydd~Jones}{2019}]%
        {guma2019we}
\bibfield{author}{\bibinfo{person}{Taulant Guma} {and} \bibinfo{person}{Rhys
  Dafydd~Jones}.} \bibinfo{year}{2019}\natexlab{}.
\newblock \showarticletitle{“Where are we going to go now?” European Union
  migrants' experiences of hostility, anxiety, and (non-) belonging during
  Brexit}.
\newblock \bibinfo{journal}{\emph{Population, Space and Place}}
  \bibinfo{volume}{25}, \bibinfo{number}{1} (\bibinfo{year}{2019}),
  \bibinfo{pages}{e2198}.
\newblock


\bibitem[\protect\citeauthoryear{Guntuku, Yaden, Kern, Ungar, and
  Eichstaedt}{Guntuku et~al\mbox{.}}{2017}]%
        {guntuku2017detecting}
\bibfield{author}{\bibinfo{person}{Sharath~Chandra Guntuku},
  \bibinfo{person}{David~B Yaden}, \bibinfo{person}{Margaret~L Kern},
  \bibinfo{person}{Lyle~H Ungar}, {and} \bibinfo{person}{Johannes~C
  Eichstaedt}.} \bibinfo{year}{2017}\natexlab{}.
\newblock \showarticletitle{Detecting depression and mental illness on social
  media: an integrative review}.
\newblock \bibinfo{journal}{\emph{Current Opinion in Behavioral Sciences}}
  \bibinfo{volume}{18} (\bibinfo{year}{2017}), \bibinfo{pages}{43--49}.
\newblock


\bibitem[\protect\citeauthoryear{Hernandez-Suarez, Sanchez-Perez,
  Toscano-Medina, Martinez-Hernandez, Sanchez, and
  Perez-Meana}{Hernandez-Suarez et~al\mbox{.}}{2018}]%
        {hernandez2018api}
\bibfield{author}{\bibinfo{person}{Aldo Hernandez-Suarez},
  \bibinfo{person}{Gabriel Sanchez-Perez}, \bibinfo{person}{Karina
  Toscano-Medina}, \bibinfo{person}{Victor Martinez-Hernandez},
  \bibinfo{person}{Victor Sanchez}, {and} \bibinfo{person}{H{\'e}ctor
  Perez-Meana}.} \bibinfo{year}{2018}\natexlab{}.
\newblock \showarticletitle{A web scraping methodology for bypassing twitter
  API restrictions}.
\newblock \bibinfo{journal}{\emph{arXiv preprint arXiv:1803.09875}}
  (\bibinfo{year}{2018}).
\newblock


\bibitem[\protect\citeauthoryear{Jofre-Bonet, Serra-Sastre, and
  Vandoros}{Jofre-Bonet et~al\mbox{.}}{2018}]%
        {jofre2018impact}
\bibfield{author}{\bibinfo{person}{Mireia Jofre-Bonet},
  \bibinfo{person}{Victoria Serra-Sastre}, {and} \bibinfo{person}{Sotiris
  Vandoros}.} \bibinfo{year}{2018}\natexlab{}.
\newblock \showarticletitle{The impact of the Great Recession on health-related
  risk factors, behaviour and outcomes in England}.
\newblock \bibinfo{journal}{\emph{Social Science \& Medicine}}
  \bibinfo{volume}{197} (\bibinfo{year}{2018}), \bibinfo{pages}{213--225}.
\newblock


\bibitem[\protect\citeauthoryear{Kim, Gilbert, Edwards, and Graeff}{Kim
  et~al\mbox{.}}{2009}]%
        {kim2009detecting}
\bibfield{author}{\bibinfo{person}{Elsa Kim}, \bibinfo{person}{Sam Gilbert},
  \bibinfo{person}{Michael~J Edwards}, {and} \bibinfo{person}{Erhardt Graeff}.}
  \bibinfo{year}{2009}\natexlab{}.
\newblock \showarticletitle{Detecting sadness in 140 characters: Sentiment
  analysis and mourning Michael Jackson on Twitter}.
\newblock \bibinfo{journal}{\emph{Web Ecology}}  \bibinfo{volume}{3}
  (\bibinfo{year}{2009}), \bibinfo{pages}{1--15}.
\newblock


\bibitem[\protect\citeauthoryear{Lampos, Lansdall-Welfare, Araya, and
  Cristianini}{Lampos et~al\mbox{.}}{2013}]%
        {lampos2013analysing}
\bibfield{author}{\bibinfo{person}{Vasileios Lampos}, \bibinfo{person}{Thomas
  Lansdall-Welfare}, \bibinfo{person}{Ricardo Araya}, {and}
  \bibinfo{person}{Nello Cristianini}.} \bibinfo{year}{2013}\natexlab{}.
\newblock \showarticletitle{Analysing mood patterns in the United Kingdom
  through Twitter content}.
\newblock \bibinfo{journal}{\emph{arXiv preprint arXiv:1304.5507}}
  (\bibinfo{year}{2013}).
\newblock


\bibitem[\protect\citeauthoryear{Lansdall-Welfare, Dzogang, and
  Cristianini}{Lansdall-Welfare et~al\mbox{.}}{2016}]%
        {lansdall2016change}
\bibfield{author}{\bibinfo{person}{Thomas Lansdall-Welfare},
  \bibinfo{person}{Fabon Dzogang}, {and} \bibinfo{person}{Nello Cristianini}.}
  \bibinfo{year}{2016}\natexlab{}.
\newblock \showarticletitle{Change-point analysis of the public mood in UK
  Twitter during the Brexit referendum}. In \bibinfo{booktitle}{\emph{2016 IEEE
  16th International Conference on Data Mining Workshops (ICDMW)}}. IEEE,
  \bibinfo{pages}{434--439}.
\newblock


\bibitem[\protect\citeauthoryear{Lantz}{Lantz}{2013}]%
        {lantz2013machine}
\bibfield{author}{\bibinfo{person}{Brett Lantz}.}
  \bibinfo{year}{2013}\natexlab{}.
\newblock \bibinfo{booktitle}{\emph{Machine learning with R}}.
\newblock \bibinfo{publisher}{Packt publishing ltd}.
\newblock


\bibitem[\protect\citeauthoryear{Liu, Shi, Ji, and Jia}{Liu
  et~al\mbox{.}}{2019}]%
        {liu2019survey}
\bibfield{author}{\bibinfo{person}{Ruijun Liu}, \bibinfo{person}{Yuqian Shi},
  \bibinfo{person}{Changjiang Ji}, {and} \bibinfo{person}{Ming Jia}.}
  \bibinfo{year}{2019}\natexlab{}.
\newblock \showarticletitle{A survey of sentiment analysis based on transfer
  learning}.
\newblock \bibinfo{journal}{\emph{IEEE Access}}  \bibinfo{volume}{7}
  (\bibinfo{year}{2019}), \bibinfo{pages}{85401--85412}.
\newblock


\bibitem[\protect\citeauthoryear{Lyons}{Lyons}{2019a}]%
        {Lyons2019pro}
\bibfield{author}{\bibinfo{person}{Kate Lyons}.}
  \bibinfo{year}{2019}\natexlab{a}.
\newblock \showarticletitle{Day democracy died: What the papers say about
  proroguing parliament}.
\newblock
  \bibinfo{journal}{\emph{https://www.theguardian.com/politics/2019/aug/29/day-democracy-died-what-the-papers-say-about-proroguing-parliament}}
  (\bibinfo{year}{2019}).
\newblock


\bibitem[\protect\citeauthoryear{Lyons}{Lyons}{2019b}]%
        {lyons2019prorogue}
\bibfield{author}{\bibinfo{person}{Kate Lyons}.}
  \bibinfo{year}{2019}\natexlab{b}.
\newblock \showarticletitle{Day democracy died: what the papers say about
  proroguing parliament}.
\newblock
  \bibinfo{journal}{\emph{https://www.theguardian.com/politics/2019/aug/29/day-democracy-died-what-the-papers-say-about-proroguing-parliament}}
  (\bibinfo{year}{2019}).
\newblock


\bibitem[\protect\citeauthoryear{Middleton}{Middleton}{2019}]%
        {Middleton2019yellow}
\bibfield{author}{\bibinfo{person}{Lucy Middleton}.}
  \bibinfo{year}{2019}\natexlab{}.
\newblock \showarticletitle{Yellowhammer Brexit document warns of rise in crime
  and fresh food}.
\newblock
  \bibinfo{journal}{\emph{https://metro.co.uk/2019/09/11/operation-yellowhammer-documents-finally-published-government-10726063/}}
  (\bibinfo{year}{2019}).
\newblock


\bibitem[\protect\citeauthoryear{Morales, Scherer, and Levitan}{Morales
  et~al\mbox{.}}{2017}]%
        {morales2017cross}
\bibfield{author}{\bibinfo{person}{Michelle Morales}, \bibinfo{person}{Stefan
  Scherer}, {and} \bibinfo{person}{Rivka Levitan}.}
  \bibinfo{year}{2017}\natexlab{}.
\newblock \showarticletitle{A cross-modal review of indicators for depression
  detection systems}. In \bibinfo{booktitle}{\emph{Proceedings of the fourth
  workshop on computational linguistics and clinical psychology—From
  linguistic signal to clinical reality}}. \bibinfo{pages}{1--12}.
\newblock


\bibitem[\protect\citeauthoryear{Nguyen, Varghese, and Barker}{Nguyen
  et~al\mbox{.}}{2013}]%
        {nguyen2013royal}
\bibfield{author}{\bibinfo{person}{Vu~Dung Nguyen}, \bibinfo{person}{Blesson
  Varghese}, {and} \bibinfo{person}{Adam Barker}.}
  \bibinfo{year}{2013}\natexlab{}.
\newblock \showarticletitle{The royal birth of 2013: Analysing and visualising
  public sentiment in the uk using twitter}. In \bibinfo{booktitle}{\emph{2013
  IEEE International Conference on Big Data}}. IEEE, \bibinfo{pages}{46--54}.
\newblock


\bibitem[\protect\citeauthoryear{Pan and Yang}{Pan and Yang}{2010}]%
        {pan2010survey}
\bibfield{author}{\bibinfo{person}{Sinno~Jialin Pan} {and}
  \bibinfo{person}{Qiang Yang}.} \bibinfo{year}{2010}\natexlab{}.
\newblock \showarticletitle{A survey on transfer learning. IEEE Transactions on
  knowledge and data engineering}.
\newblock \bibinfo{journal}{\emph{22 (10): 1345}}  \bibinfo{volume}{1359}
  (\bibinfo{year}{2010}).
\newblock


\bibitem[\protect\citeauthoryear{Pennebaker, Francis, and Booth}{Pennebaker
  et~al\mbox{.}}{2001}]%
        {pennebaker2001linguistic}
\bibfield{author}{\bibinfo{person}{James~W Pennebaker},
  \bibinfo{person}{Martha~E Francis}, {and} \bibinfo{person}{Roger~J Booth}.}
  \bibinfo{year}{2001}\natexlab{}.
\newblock \showarticletitle{Linguistic inquiry and word count: LIWC 2001}.
\newblock \bibinfo{journal}{\emph{Mahway: Lawrence Erlbaum Associates}}
  \bibinfo{volume}{71}, \bibinfo{number}{2001} (\bibinfo{year}{2001}),
  \bibinfo{pages}{2001}.
\newblock


\bibitem[\protect\citeauthoryear{Shimodaira}{Shimodaira}{2000}]%
        {shimodaira2000improving}
\bibfield{author}{\bibinfo{person}{Hidetoshi Shimodaira}.}
  \bibinfo{year}{2000}\natexlab{}.
\newblock \showarticletitle{Improving predictive inference under covariate
  shift by weighting the log-likelihood function}.
\newblock \bibinfo{journal}{\emph{Journal of statistical planning and
  inference}} \bibinfo{volume}{90}, \bibinfo{number}{2} (\bibinfo{year}{2000}),
  \bibinfo{pages}{227--244}.
\newblock


\bibitem[\protect\citeauthoryear{Tumasjan, Sprenger, Sandner, and
  Welpe}{Tumasjan et~al\mbox{.}}{2010}]%
        {tumasjan2010predicting}
\bibfield{author}{\bibinfo{person}{Andranik Tumasjan}, \bibinfo{person}{Timm~O
  Sprenger}, \bibinfo{person}{Philipp~G Sandner}, {and}
  \bibinfo{person}{Isabell~M Welpe}.} \bibinfo{year}{2010}\natexlab{}.
\newblock \showarticletitle{Predicting elections with twitter: What 140
  characters reveal about political sentiment}. In
  \bibinfo{booktitle}{\emph{Fourth international AAAI conference on weblogs and
  social media}}.
\newblock


\bibitem[\protect\citeauthoryear{Vandoros, Avendano, and Kawachi}{Vandoros
  et~al\mbox{.}}{2019}]%
        {vandoros2019eu}
\bibfield{author}{\bibinfo{person}{Sotiris Vandoros}, \bibinfo{person}{Mauricio
  Avendano}, {and} \bibinfo{person}{Ichiro Kawachi}.}
  \bibinfo{year}{2019}\natexlab{}.
\newblock \showarticletitle{The EU referendum and mental health in the short
  term: a natural experiment using antidepressant prescriptions in England}.
\newblock \bibinfo{journal}{\emph{J Epidemiol Community Health}}
  \bibinfo{volume}{73}, \bibinfo{number}{2} (\bibinfo{year}{2019}),
  \bibinfo{pages}{168--175}.
\newblock


\bibitem[\protect\citeauthoryear{Weiss, Khoshgoftaar, and Wang}{Weiss
  et~al\mbox{.}}{2016}]%
        {weiss2016survey}
\bibfield{author}{\bibinfo{person}{Karl Weiss}, \bibinfo{person}{Taghi~M
  Khoshgoftaar}, {and} \bibinfo{person}{DingDing Wang}.}
  \bibinfo{year}{2016}\natexlab{}.
\newblock \showarticletitle{A survey of transfer learning}.
\newblock \bibinfo{journal}{\emph{Journal of Big data}} \bibinfo{volume}{3},
  \bibinfo{number}{1} (\bibinfo{year}{2016}), \bibinfo{pages}{9}.
\newblock


\bibitem[\protect\citeauthoryear{Wongkoblap, Vadillo, and Curcin}{Wongkoblap
  et~al\mbox{.}}{2017}]%
        {wongkoblap2017researching}
\bibfield{author}{\bibinfo{person}{Akkapon Wongkoblap},
  \bibinfo{person}{Miguel~A Vadillo}, {and} \bibinfo{person}{Vasa Curcin}.}
  \bibinfo{year}{2017}\natexlab{}.
\newblock \showarticletitle{Researching mental health disorders in the era of
  social media: systematic review}.
\newblock \bibinfo{journal}{\emph{Journal of medical Internet research}}
  \bibinfo{volume}{19}, \bibinfo{number}{6} (\bibinfo{year}{2017}),
  \bibinfo{pages}{e228}.
\newblock


\bibitem[\protect\citeauthoryear{Yazdavar, Mahdavinejad, Bajaj, Romine, Sheth,
  Monadjemi, Thirunarayan, Meddar, Myers, Pathak, et~al\mbox{.}}{Yazdavar
  et~al\mbox{.}}{2020}]%
        {yazdavar2020multimodal}
\bibfield{author}{\bibinfo{person}{Amir~Hossein Yazdavar},
  \bibinfo{person}{Mohammad~Saeid Mahdavinejad}, \bibinfo{person}{Goonmeet
  Bajaj}, \bibinfo{person}{William Romine}, \bibinfo{person}{Amit Sheth},
  \bibinfo{person}{Amir~Hassan Monadjemi}, \bibinfo{person}{Krishnaprasad
  Thirunarayan}, \bibinfo{person}{John~M Meddar}, \bibinfo{person}{Annie
  Myers}, \bibinfo{person}{Jyotishman Pathak}, {et~al\mbox{.}}}
  \bibinfo{year}{2020}\natexlab{}.
\newblock \showarticletitle{Multimodal mental health analysis in social media}.
\newblock \bibinfo{journal}{\emph{Plos one}} \bibinfo{volume}{15},
  \bibinfo{number}{4} (\bibinfo{year}{2020}), \bibinfo{pages}{e0226248}.
\newblock


\bibitem[\protect\citeauthoryear{Zhang, Zhang, Chan, and Rosso}{Zhang
  et~al\mbox{.}}{2019}]%
        {zhangairony}
\bibfield{author}{\bibinfo{person}{Shiwei Zhang}, \bibinfo{person}{Xiuzhen
  Zhang}, \bibinfo{person}{Jeffrey Chan}, {and} \bibinfo{person}{Paolo Rosso}.}
  \bibinfo{year}{2019}\natexlab{}.
\newblock \showarticletitle{Irony detection via sentiment-based transfer
  learning}.
\newblock \bibinfo{journal}{\emph{Information Processing \& Management}}
  \bibinfo{volume}{56}, \bibinfo{number}{5} (\bibinfo{year}{2019}),
  \bibinfo{pages}{1633--1644}.
\newblock


\end{thebibliography}

\end{document}